\pdfoutput=1

\documentclass[11pt]{article}
\usepackage[]{emnlp}
\usepackage{times}
\usepackage{latexsym}
\usepackage[T1]{fontenc}
\usepackage[utf8]{inputenc}
\usepackage{microtype}

\usepackage{epsfig}
\usepackage{graphicx}
\usepackage{amsmath}
\usepackage{amssymb}
\usepackage{url}
\usepackage{booktabs}
\usepackage{subfigure}
\usepackage{bm}
\usepackage{color}
\usepackage{pifont}
\usepackage{amsfonts}
\usepackage{stmaryrd}
\usepackage{stfloats}
\usepackage{xspace}
\usepackage{threeparttable}
\usepackage{adjustbox}
\usepackage{algorithm,algorithmicx}
\usepackage{algpseudocode}
\usepackage{multirow}
\usepackage{soul}
\usepackage{ragged2e}
\usepackage{enumitem}

\newcommand{\ie}{\emph{i.e.,}\xspace}
\newcommand{\eg}{\emph{e.g.,}\xspace}

\title{Towards Debiasing Temporal Sentence Grounding in Video}

\author{
   Hao Zhang$^{1,2}$,~~Aixin Sun$^{1}$,~~Wei Jing$^{3}$,~~Joey Tianyi Zhou$^{2}$ \\
   $^{1}$School of Computer Science and Engineering, Nanyang Technological University, Singapore \\
   $^{2}$Institute of High Performance Computing, A*STAR, Singapore \\
   $^{3}$Alibaba DAMO Academy, China \\
   \texttt{hao007@e.ntu.edu.sg,}~~
   \texttt{axsun@ntu.edu.sg} \\
   \texttt{21wjing@gmail.com,}~~
   \texttt{joey\_zhou@ihpc.a-star.edu.sg}
}

\begin{document}
\maketitle
\begin{abstract}
The temporal sentence grounding in video (TSGV) task is to locate a temporal moment from an untrimmed video, to match a language query, \ie a sentence. Without considering bias in moment annotations (\eg start and end positions in a video), many models tend to capture statistical regularities of the moment annotations, and do not well learn cross-modal reasoning between video and language query. In this paper, we propose two debiasing strategies, \textit{data debiasing} and \textit{model debiasing}, to ``force'' a TSGV model to capture cross-modal interactions. Data debiasing performs data oversampling through video truncation to balance moment temporal distribution in train set. Model debiasing leverages video-only and query-only models to capture the distribution bias, and forces the model to learn  cross-modal interactions. Using VSLNet as the base model, we evaluate impact of the two strategies on two datasets that contain out-of-distribution test instances. Results show that both strategies are effective in improving model generalization capability. Equipped with both debiasing strategies, VSLNet achieves best results on both datasets.
\end{abstract}

\section{Introduction}\label{sec:intro}
The goal of temporal sentence grounding in video (TSGV)  is to retrieve the temporal moment that matches a language query, from an untrimmed video. As a fundamental yet challenging task in the vision-language understanding area, TSGV attracts increasing attentions. A good number of models have been proposed recently~\cite{Gao2017TALLTA,chen2018temporally,yuan2019semantic,chen2020rethinking,xiao2021boundary,zhang2021video}. In general, a model needs to have deep understanding of the video, the language query, and their cross-modal interactions to make accurate predictions.

\begin{figure}[t]
    \centering
	\subfigure[\small Charades-CD]
	{\label{fig:charades_cd_dist}	\includegraphics[width=0.47\textwidth]{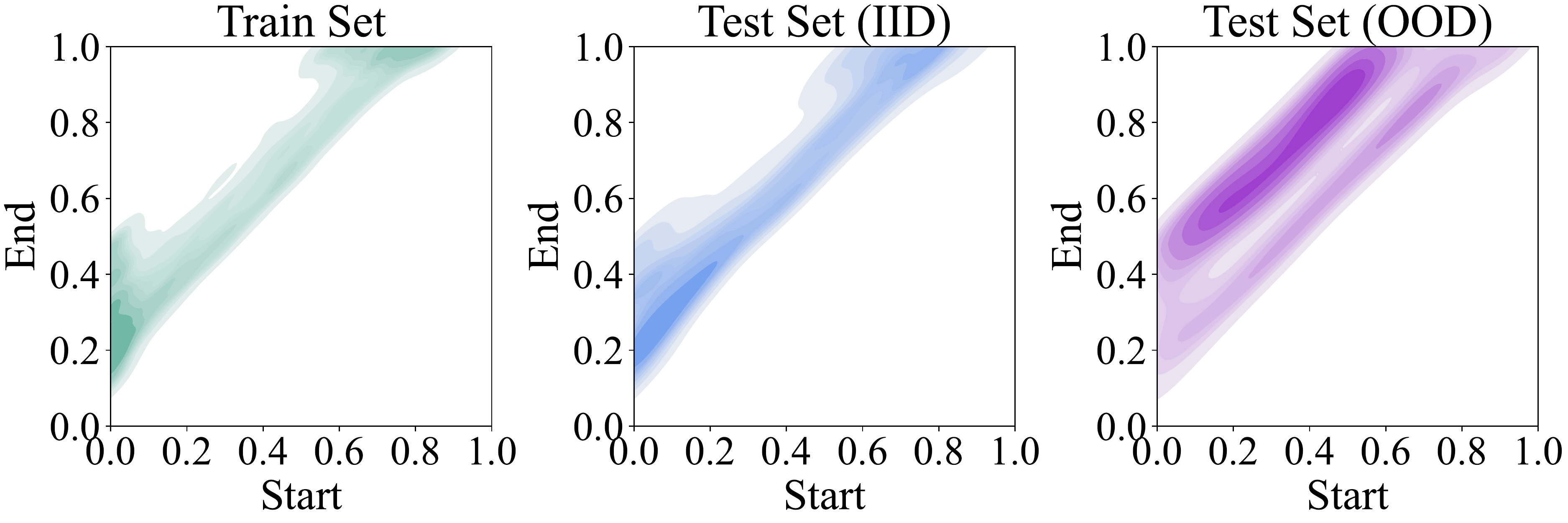}}
	\subfigure[\small ActivityNet-CD]
	{\label{fig:activitynet_cd_dist}	\includegraphics[width=0.47\textwidth]{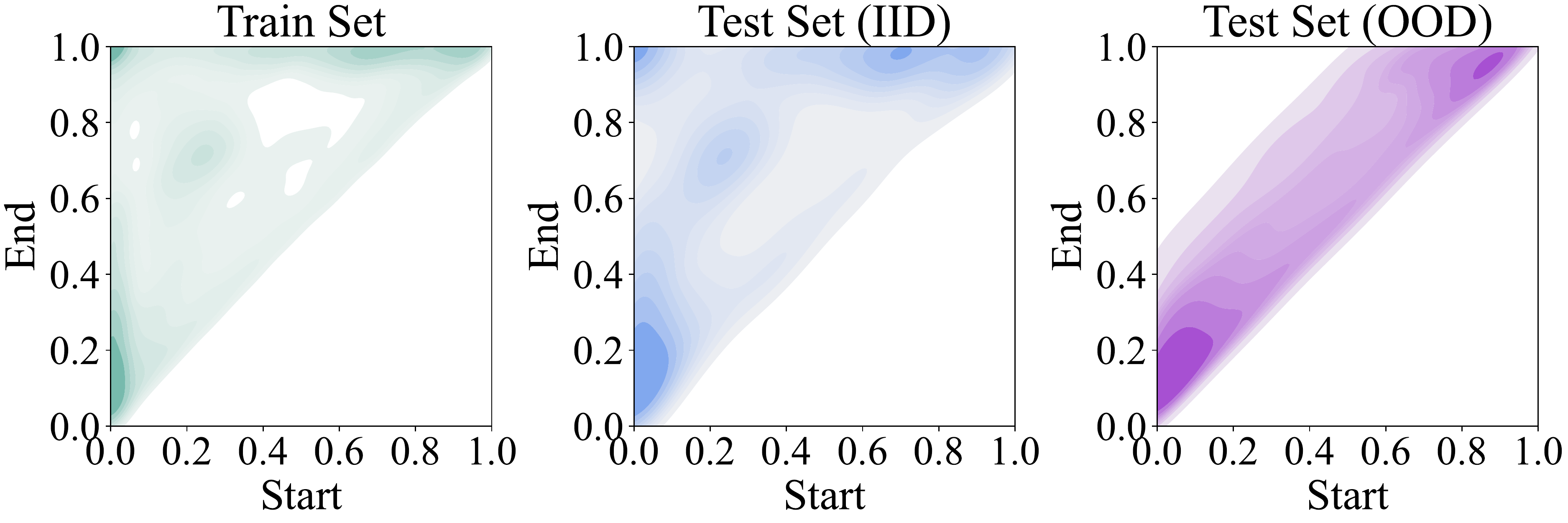}}
	\caption{\small Moment annotation distributions of Charades-CD and ActivityNet-CD datasets~\citet{yuan2021closer}, where ``Start'' and ``End'' axes represent the normalized start and end time points, respectively. Deeper color represents larger density (\ie more annotations) in the dataset.}
	\label{fig:cd_dist}
\end{figure}

Recent studies~\cite{otani2020challengesmr,yuan2021closer} noted that 1) substantial \textit{distribution bias} exists in benchmark datasets; 2) many TSGV models rely on exploiting the statistical regularities of annotation distribution for moment retrieval, to achieve good performance. We use the train set of Charades-CD dataset as an example to illustrate distribution bias, shown in Figure~\ref{fig:charades_cd_dist}. The ``Start'' and ``End'' axes denote the normalized start and end time points of the annotated moments in videos. Observe that many annotated moments in the train set of Charades-CD locate in the normalized temporal region of $0.2\sim 0.4$ of the video length. Hence, a TSGV model could make a good guess of start/end time points, even without taking into consideration the input video and language query. Consequently, the existing TSGV methods achieve impressive performance on the independent-and-identical distribution (iid) test set, but fail to generalize on the out-of-distribution (ood) test set (see Figure~\ref{fig:cd_dist}). For instance, without debasing, the VSLNet model reports $39.25\%$ on the iid test set of Charades-CD, but only $27.20\%$ on the ood test set of the same dataset (see Section~\ref{sec:experiment} for details). 

The aforementioned studies~\cite{otani2020challengesmr,yuan2021closer} well analyze the issue of distribution bias in TSGV datasets and models. However, they do not attempt to address the bias. To the best of our knowledge, this work is the first attempt to mitigate bias in TSGV models. Specifically, we propose two debiasing strategies, data debiasing and model debiasing.

From data perspective, bias is caused by the imbalanced distribution of moment annotations, \ie many annotations are concentrated in several regions as shown in Figure~\ref{fig:cd_dist}. 
To mitigate the bias, we artificially re-balance the moment annotations via data augmentation, \ie creating more samples through video truncation. Specifically, we partition each training video into multiple non-overlapping clips, and gradually cut background clips (\ie clips that do not contain target moment) to form new videos. Each newly created video hence has a different start/end time point from the original training sample. Besides, each query corresponds to multiple videos of different lengths after data debiasing. This oversampling implicitly encourages a TSGV model to focus more on the interactions between query and target moment.

Our model debiasing strategy is inspired by the question-only branch in visual question answering (VQA) debiasing~\cite{cadene2019rubi,clark2019dont}. Bias in VQA comes from the suspicious correlations between answer occurrences and certain patterns of questions, \eg color of a banana is always answered as ``yellow'' regardless the input image contains a yellow or green banana. Bias in TSGV, on the other hand, is caused by the regularities in the moment boundaries. That is, a TSGV model with video- or query-only input could achieve fairly good performance by making a good guess of the distribution bias. 
Thus, our model debiasing is different from the strategies in VQA models. To debias, we add two unimodal models, \ie video- and query-only branches, in addition to the TSGV model. Both unimodal models learn to capture the bias, and to disentangle bias from the TSGV model by adjusting losses to compensate for biases dynamically. Specifically, the gradients backpropagated through TSGV model are reduced for the biased examples and are increased for the less biased after loss adjustment.

We evaluate the proposed data and model debiasing strategies on the Charades-CD and ActivityNet-CD datasets, by using VSLNet as the base model. Both datasets are re-designed by~\citet{yuan2021closer} for the purpose of evaluating model biasing, with dedicated iid test set and ood test set (see Figure~\ref{fig:cd_dist}). Experimental results demonstrate their ability of well generalization on  out-of-distribution test set. With both debiasing strategies, VSLNet achieves further improvements against the base model. 

Our main contributions are: 1) to the best of our knowledge, we are the first to address distribution bias in TSGV task; 2) we propose two simple yet effective strategies, data debiasing (DD) and model debiasing (MD), to alleviate the bias issue from the data level and model level, respectively; 3) we conduct extensive experiments on benchmark datasets to demonstrate that both strategies effectively reduce the bias in TSGV base model, and also lead to further overall performance improvement.

\section{Related Work}\label{sec:related}
\paragraph{Temporal sentence grounding in video} Solutions to TSGV can be roughly categorized into proposal-based, proposal-free, reinforcement learning, and weakly supervised methods. 

\textit{Proposal-based} methods include ranking and anchor-based models. Ranking-based models~\cite{Gao2017TALLTA,hendricks2017localizing,hendricks2018localizingM,Liu2018CML,chen2019semantic,ge2019mac,Xu2019MultilevelLA,xiao2021boundary} solve TSGV with a propose-and-rank pipeline. Proposals (short video clips) are first generated, then the best ranked  proposal for a query is retrieved via multimodal matching. Anchor-based models~\cite{chen2018temporally,yuan2019semantic,zhang2019man,zhu2019cross,Wang2020TemporallyGL} solve TSGV in an end-to-end manner. The model sequentially scores multiscale anchors ended at each frame and generates results in a single pass. 

\textit{Proposal-free} methods~\cite{lu2019debug,Yuan2019ToFW,ghosh2019excl,chen2020rethinking,zhang2020learning,zeng2020dense,li2021proposal,zhang2021qa4nlvl,zhang2021parallel,nan2021interventional} tackle TSGV by learning fine-grained cross-modal interactions for video and query. The boundaries of target moment are then predicted or regressed  directly. 
\textit{Reinforcement learning} methods~\cite{he2019Readwa,Wang2019LanguageDrivenTA,hahn2020tripping,Wu2020TreeStructuredPB} formulate TSGV as a sequential decision-making problem. These methods imitate humans' coarse-to-fine decision-making process. Adopting reinforcement learning algorithms, the models progressively observe candidate moments conditioned on the query. 
\textit{Weakly supervised} methods~\cite{duan2018weakly,mithun2019weakly,gao2019wslln,song2020weakly,Lin2020WeaklySupervisedVM,tan2021logan} consider the ground truth moments are unavailable in the training stage. These methods explore to learn the latent alignment between video content and language query, and retrieve the target moment based on their similarities.

\paragraph{Bias in TSGV task} 
The unimodal bias has been identified and well studied in VQA task~\cite{Goyal2017MakingTV,agrawal2018dont}. This bias is considered as the main issue for the generalization ability of VQA models.
To remedy, adversary-based~\cite{ramakrishnan2018overcoming,grand2019adversarial} and fusion-based~\cite{cadene2019rubi,clark2019dont,Chen2020Counterfactual,liang2020learning} methods are proposed to mitigate the bias in VQA and make great progress. 

Inspired by the studies on biases in VQA, \citet{otani2020challengesmr} first assessed how well benchmark results reflect the true progress in solving TSGV. They developed several bias-based models and evaluated their performances on TSGV datasets. Their results show substantial biases in TSGV datasets, and SOTA methods tend to fit on distribution bias to achieve good performance rather than learning cross-modal reasoning between video and query.  \citet{yuan2021closer} further analyze the moment annotation distributions of Charades-STA~\cite{Gao2017TALLTA} and ActivityNet Captions~\cite{krishna2017dense} datasets, and show the distributions of their train and test sets are almost identical. The authors then develop Charades-CD and ActivityNet-CD datasets, where each dataset consists of train, validation, iid and ood test sets. The performance gap between the iid and ood test sets  effectively reflects the generalization ability of the TSGV model. Note that, both studies~\cite{otani2020challengesmr,yuan2021closer} aim to offer deep understanding on bias in TSGV, and they do not focus on debiasing models. A recent work~\cite{zhou2021embracing} proposes to solve  bias TSGV. However, it only aims to alleviate the bias caused by single-style annotations in the presence of label uncertainty. In contrast, our work focuses on addressing the annotation distribution bias of TSGV task.

\section{Data Debiasing and Model Debiasing}\label{sec:method}
We first define and formulate the TSGV task in feature space. Then we elaborate the proposed data and model debiasing strategies. The data debiasing strategy is model-agnostic, hence can be applied to any TSGV model in principle. The proposed model debiasing strategy is applicable to proposal-free models. Lastly, we use VSLNet~\cite{zhang2020vslnet} as a backbone to illustrate how to apply the two debiasing strategies.

\subsection{Preliminaries}\label{ssec:pre}
We denote an untrimmed video with $T$ frames as $F=[f_t]_{t=0}^{T-1}$, language query with $the n_q$ words as $Q=[q_i]_{i=0}^{n_q-1}$, $t^s$ and $t^e$ as the start and end time of ground truth moment. The video $V$ is then split into $n_v$ units and encoded into visual features $\bm{V}=[\bm{v}_i]_{i=0}^{n_v-1}\in\mathbb{R}^{n_v\times d_v}$ with pre-trained feature extractor. The query $Q$ is initialized by pre-trained word embeddings as $\bm{Q}=[\bm{w}_i]_{i=0}^{n_q-1}\in\mathbb{R}^{n_q\times d_v}$. The $t^{s(e)}$ are mapped to the corresponding indices $i^{s(e)}$ in the feature sequence, where $0 \leq i^s \leq i^e \leq n_v-1$. The goal of TSGV is to localize the moment starting at $i^s$ and ending at $i^e$.

\begin{figure}[t]
    \centering
	\subfigure[\small An illustration of the data debiasing strategy.]
	{\label{fig:data_debiasing_strategy}	\includegraphics[width=0.23\textwidth]{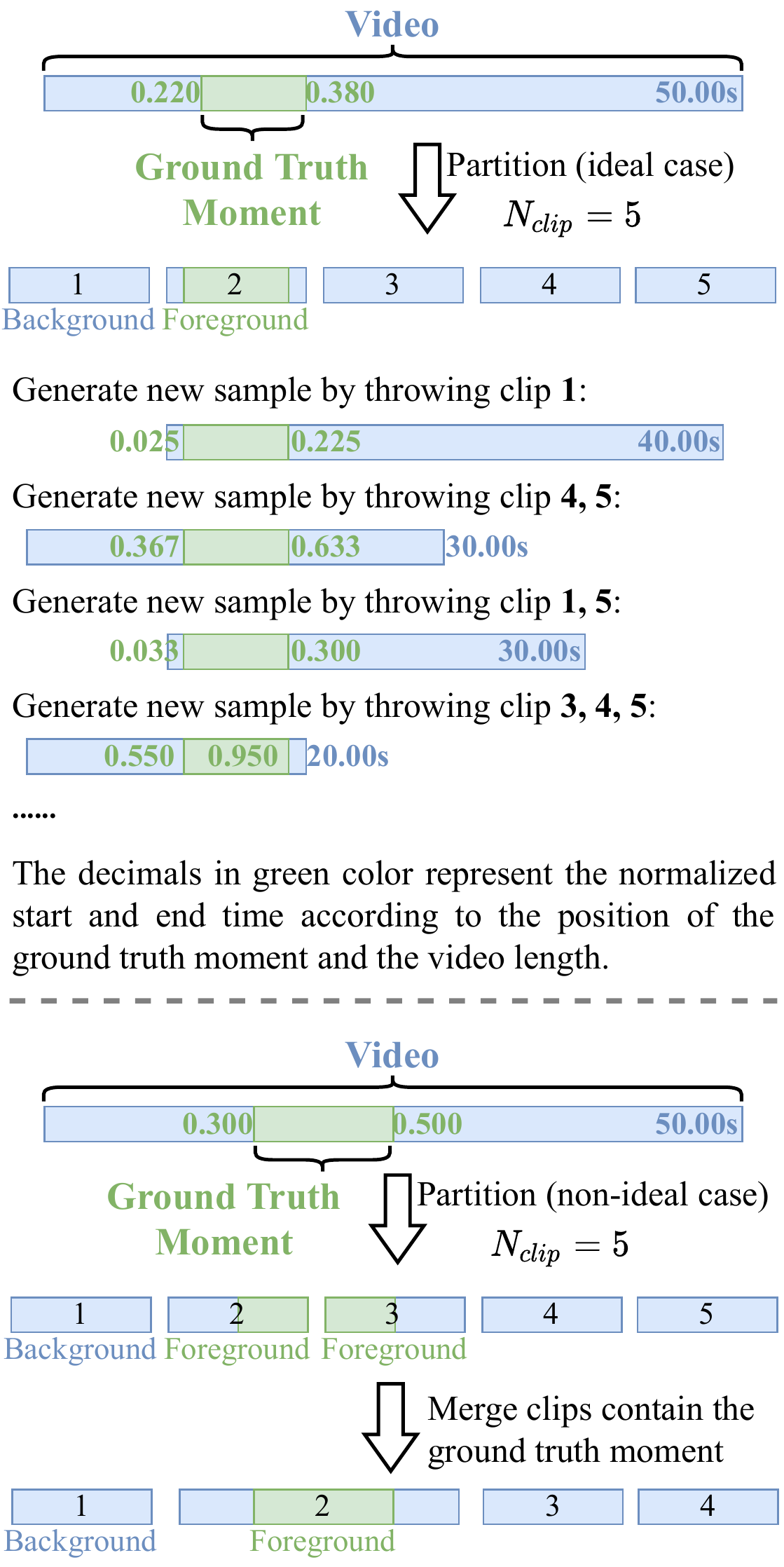}}
	\subfigure[\small The train set distribution before/after data debiasing.]
	{\label{fig:charades_cd_data_debiasing}	\includegraphics[width=0.23\textwidth]{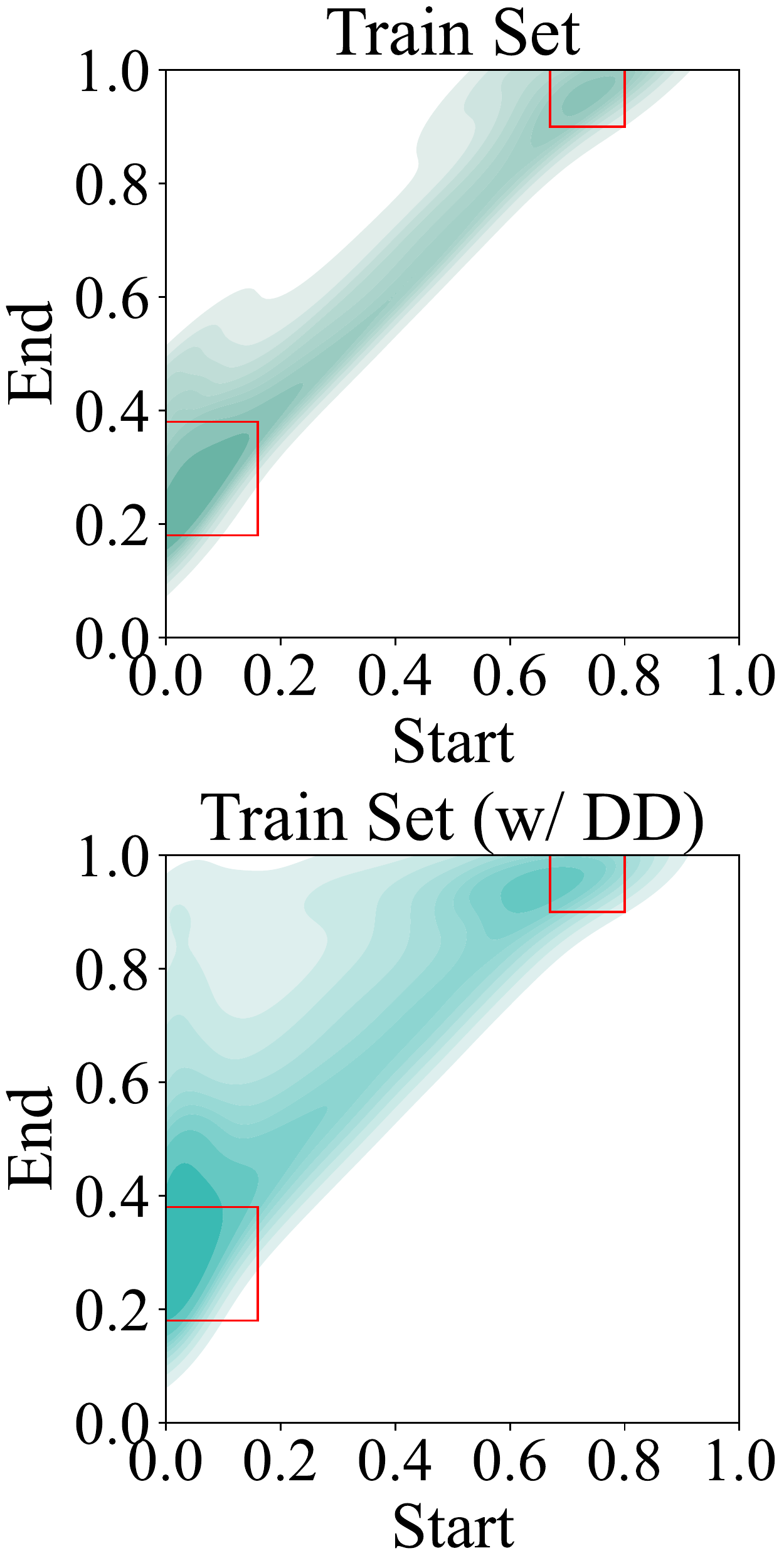}}
	\caption{\small The illustration of data debiasing (DD) strategy and an example of augmented train set distribution for Charades-CD dataset. Note the red rectangles in (b) highlight the regions of biased samples.}
	\label{fig:data_debiasing}
\end{figure}

\subsection{Data Debiasing}\label{ssec:dd}
As shown in Figure~\ref{fig:cd_dist}, the iid test set shares identical distribution of start/end positions with the train set, while ood test has an entirely different distribution. From data perspective, this issue is caused by the fact that train set does not cover many moment annotations that start and end at different positions of the untrimmed video. Thus, we propose a simple data debiasing (DD) strategy to include more annotations with varying start/end positions. 

Ideally, the moment annotations of train set should uniformly distribute in the upper triangle area in the distribution plot (see Figure~\ref{fig:cd_dist}). To this end, we oversample annotations through video truncation. Illustrated in Figure~\ref{fig:data_debiasing_strategy}, for each video-query pair, we first partition the video into multiple non-overlapping clips. If a clip overlaps with ground truth moment, then it is regarded as foreground, otherwise as background. If the ground truth moment happens to be partitioned into multiple clips, then these clips are merged back to ensure the ground truth moment is unaffected. Then we gradually truncate the video to generate new videos by throwing background clips from both ends. The newly generated videos hence are sub-clips of the original video, with their overall length reduced, but all contain the ground truth moment. The relative positions of start and end boundaries of ground truth moment are then well distributed. Because the correspondence between language query and ground truth moment does not change, this oversampling process makes each query correspond to multiple videos of different lengths. In this way, data debiasing implicitly encourages  TSGV model to learn the multimodal matching between query and the target moment, with different amount of irrelevant content in videos. Note that, we do not rotate or permutate background clips, to ensure the semantic continuity in the generated videos.

We use the train set of Charades-CD as an example to illustrate the effect of DD in Figure~\ref{fig:charades_cd_data_debiasing}. After data debiasing, moment annotation distribution of the augmented train set spreads to a much larger area than that of the original train set. Besides, the proportion of biased samples (highlighted in red rectangle in the figure) in the entire training samples reduces from $38.83\%$ to $21.80\%$. 

\begin{figure}[t]
    \centering
    \includegraphics[trim={0cm 0cm 2.9cm 0cm},clip,width=0.48\textwidth]{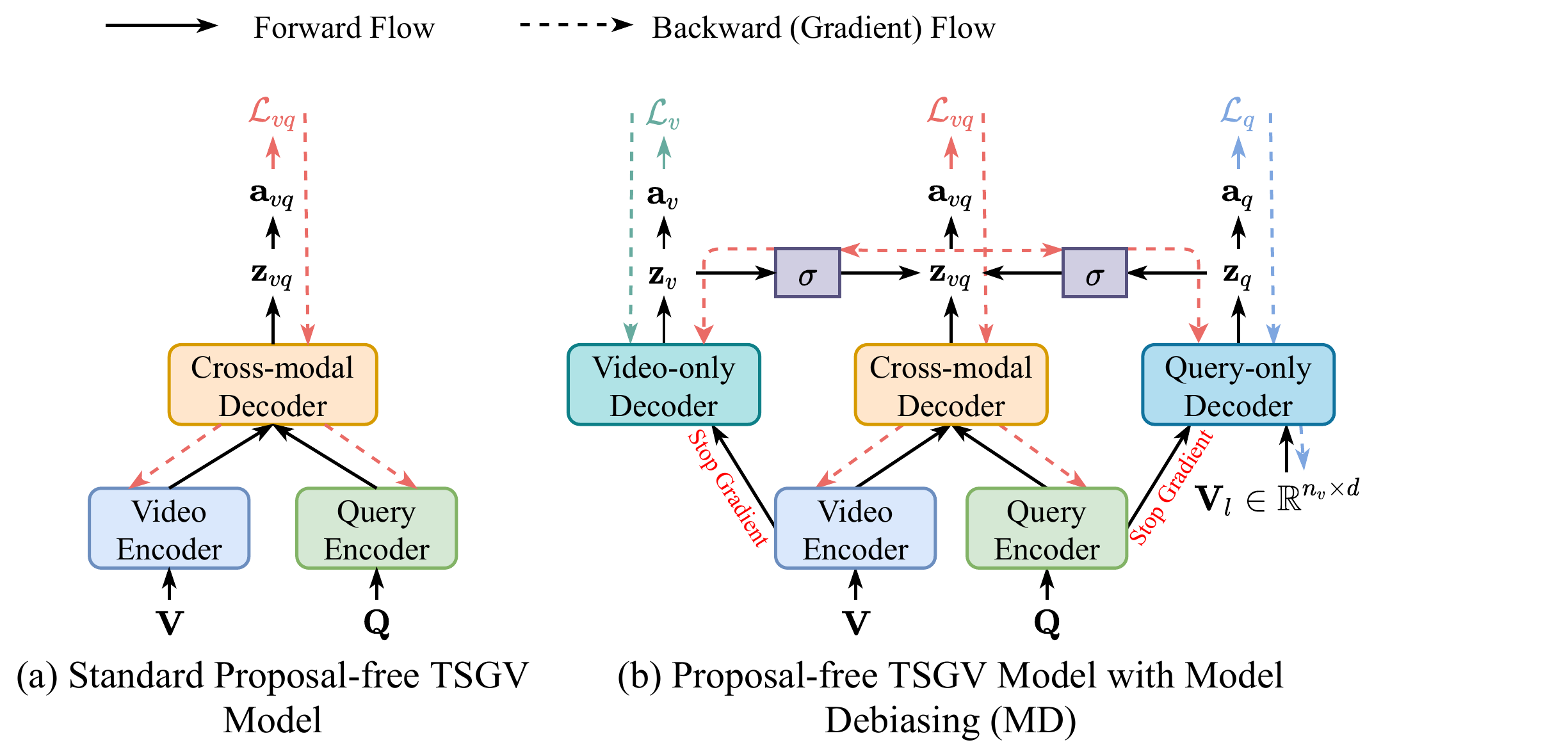}
    \caption{\small A standard proposal-free TSGV model (a), and model debiasing strategy (b).}
	\label{fig:model_debiasing}
\end{figure}

\subsection{Model Debiasing}\label{ssec:md}
We propose model debiasing for \textit{proposal-free} TSGV models, which directly learn cross-modal interactions between video and query, and directly predict boundaries of target moment. Next, we use the computation process of span-based proposal-free TSGV model to present our debiasing strategy; the strategy can be applied to regression-based model with minor adaptations. 

A span-based model usually consists of a video encoder $e_v:\bm{V}\in\mathbb{R}^{n_v\times d_v}\rightarrow\bm{\bar{V}}\in\mathbb{R}^{n_v\times d}$, a query encoder $e_q:\bm{Q}\in\mathbb{R}^{n_q\times d_q}\rightarrow\bm{\bar{Q}}\in\mathbb{R}^{n_q\times d}$, a cross-modal interaction module $m_{vq}:\bm{\bar{V}}\times\bm{\bar{Q}}\rightarrow\bm{H}_{vq}\in\mathbb{R}^{n_v\times d}$, and an answer predictor $g_{vq}:\bm{H}_{vq}\rightarrow \bm{z}_{vq}=\{\bm{z}_{vq}^s, \bm{z}_{vq}^e\}\in\mathbb{R}^{n_v\times 2}$. The overall architecture is shown in Figure~\ref{fig:model_debiasing}(a), where we combine the cross-modal interaction module $m_{vq}$ and answer predictor $g_{vq}$ as cross-modal decoder $\varphi_{vq}:\bm{\bar{V}}\times\bm{\bar{Q}}\rightarrow \bm{z}_{vq}=\{\bm{z}_{vq}^s, \bm{z}_{vq}^e\}\in\mathbb{R}^{n_v\times 2}$ for better visualization, and $\bm{a}_{vq}^{s(e)}=\mathtt{Softmax}(\bm{z}_{vq}^{s(e)})$. The training objective is defined as:
\begin{equation}
    \mathcal{L}_{\text{vq}} = \frac{1}{2}\times\big[f_{\text{XE}}(\bm{a}_{vq}^s, \bm{y}^s) + f_{\text{XE}}(\bm{a}_{vq}^e, \bm{y}^e)\big]
\label{eq:vq_loss}
\end{equation}
where $f_{\text{XE}}$ is cross-entropy function, $\bm{y}^{s(e)}$ is one-hot labels for start (end) boundary, \ie $i^s$($i^e$).

Due to the existence of substantial data distribution biases in TSGV datasets,  TSGV models tend to rely on the statistical regularities  to generate predictions even without having to consider the video and query inputs~\cite{otani2020challengesmr}. Hence, a TSGV model with unimodal input (\ie either video or query) could also achieve fair performance by capturing the distribution biases in TSGV datasets. Inspired by~\cite{cadene2019rubi,clark2019dont}, we adapt unimodal models as separate branches and integrate them into TSGV model, illustrated in Figure~\ref{fig:model_debiasing}(b). The unimodal models learn to capture the distribution bias, and force the TSGV model to focus on learning cross-modal interaction between video and query by altering its answer predictions. 

Specifically, given the encoded visual ($\bm{\bar{V}}$) and textual ($\bm{\bar{Q}}$) features, the TSGV model retrieves the answer with cross-modal decoder as:
\begin{equation}
    \bm{z}_{vq} = g_{vq}(m_{vq}(\bm{\bar{V}}, \bm{\bar{Q}}))
\end{equation}
Because the video-only branch does not contain query input, the video-only decoder generate answer by:
\begin{equation}
    \bm{z}_v = g_v(\bm{\bar{V}})
\end{equation}
For query-only branch, as TSGV task requires to retrieve  start and end boundaries of target moment, we follow~\citet{otani2020challengesmr} to replace the $\bm{\bar{V}}$ with a learnable feature sequence $\bm{V}_l\in\mathbb{R}^{n_v\times d}$ in the same shape, to simulate visual input. The answer is computed as:
\begin{equation}
    \bm{z}_q = g_q(m_q(\bm{V}_l, \bm{\bar{Q}}))
\end{equation}
Note that, the structures of $m_q$ and $m_{vq}$ are the same, but their parameters are not shared. Also, $g_v$, $g_q$, and $g_{vq}$ are the same.

With biased prediction $\bm{z}_q$ and $\bm{z}_v$, we modify the prediction of TSGV model $\bm{z}_{vq}$ as:
\begin{equation}
    \bm{z}'_{vq} = \bm{z}_{vq} \odot \sigma(\bm{z}_q) \odot \sigma(\bm{z}_v)
\label{eq:modified_prediction}
\end{equation}
where $\odot$ denotes element-wise multiplication and $\sigma$ is Sigmoid activation to map the biased prediction between $0$ and $1$. The key of Eq.~\ref{eq:modified_prediction} is to dynamically alter the loss, by modifying  predictions of the TSGV model, to prevent the model to pick up distribution biases from the dataset.

Given a biased sample, both unimodal branches and the TSGV model tend to generate high confidence to the correct answer and low confidence to others. The confidence score of the correct answer predicted by the TSGV model will be further increased with Eq.~\ref{eq:modified_prediction}. Thus, the loss from a biased sample is much smaller. Accordingly, the gradient backpropagated through the TSGV model is very small, reducing the importance of this biased sample in training. On the contrary, the importance of a non-biased sample will be increased. The reason is, all the models tend to assign low confidence to the correct answer, hence the score is further decreased through Eq.~\ref{eq:modified_prediction}. With a large loss, the TSGV model is forced on learning non-biased samples.

Finally, we jointly optimize the two unimodal models and the TSGV model as:
\begin{equation}
    \mathcal{L}_{all} = \mathcal{L}_{vq} + \mathcal{L}_q + \mathcal{L}_v
\end{equation}
Note that, both $\mathcal{L}_q$ and $\mathcal{L}_v$ do not optimize the query ($e_q$) and video ($e_v$) encoders to prevent them from directly learning distribution biases. $\mathcal{L}_{vq}$ is used to optimize all the modules except the learnable feature sequence for query-only decoder. During inference, only the TSGV model is used.

\subsection{Implementation}\label{ssec:impl}
We use VSLNet~\cite{zhang2020vslnet} as the backbone model to evaluate both debiasing strategies, due to its simple architecture and prominent performance. VSLNet consists of a transformer-based module as video encoder $e_v$, a transformer-based module as query encoder $e_q$, a video-query co-attention layer as cross-modal reasoning module $m_{vq}$, and the stacked LSTMs as answer predictor $g_{vq}$. Here, we replace the stacked LSTMs with stacked transformer blocks~\cite{vaswani2017attention} for faster training and inference. 

In our implementation, the VSLNet is regarded as the standard TSGV model. Then we initialize a new answer predictor $g_v$ with stacked transformer blocks as video-only decoder. The query-only decoder is constructed by creating a new cross-modal reasoning module $m_q$ and a new answer predictor $g_q$. As mentioned before, the $m_q$ and $m_{vq}$ have the same structure but their parameters are not shared, while $g_v$, $g_q$ and $g_{vq}$ are the same. Finally, the VSLNet, query-only decoder and video-only decoder are integrated together following Figure~\ref{fig:model_debiasing}(b).

\section{Experiments}\label{sec:experiment}

\subsection{Experimental Setting}
\paragraph{Datasets} We conduct experiments on Charades-CD (\textbf{Cha-CD}) and ActivityNet-CD (\textbf{ANet-CD}) datasets, prepared by~\cite{yuan2021closer}. The two datasets originate from Charades-STA~\cite{Gao2017TALLTA} (\textbf{Cha-STA}) and ActivityNet Captions~\cite{krishna2017dense} (\textbf{ANetCap}) datasets, respectively. The train/test instances are reorganised by~\cite{yuan2021closer} into train, val, iid test and ood test sets. Robustness of video grounding models are measured by the performance gap between their iid and ood test sets. We also conduct experiments on Cha-STA and ANetCap to further validate our strategies. Statistics of the datasets are summarized in Table~\ref{tab:data}.

\begin{table}[t]
    \small
	\centering
	\setlength{\tabcolsep}{2.5 pt}
	\begin{tabular}{c | c c c}
		\specialrule{.1em}{.05em}{.05em}
		Dataset & Split & \# Videos & \# Annotations \\
		\specialrule{.1em}{.05em}{.05em}
		\multirow{4}{*}{Charades-CD}      & train          & 4,564  & 11,071 \\
		                                  & val            & 333    & 859    \\
		                                  & test-iid (iid) & 333    & 823    \\
		                                  & test-ood (ood) & 1,442  & 3,375  \\
		\hline
		\multirow{4}{*}{ActivityNet-CD}   & train          & 10,984 & 51,415 \\
		                                  & val            & 746    & 3,521  \\
		                                  & test-iid (iid) & 746    & 3,443  \\
		                                  & test-ood (ood) & 2,450  & 13,578 \\
		\hline
		\multirow{2}{*}{Charades-STA}     & train          & 5,338  & 12,408 \\
		                                  & test           & 1,334  & 3,720  \\
		\hline
		\multirow{2}{*}{ActivityNet Caps} & train          & 10,009 & 37,421 \\
		                                  & test           &  4,917 & 34,536 \\
        \specialrule{.1em}{.05em}{.05em}
	\end{tabular}
	\caption{\small Statistics of the TSGV benchmark datasets.}
	\label{tab:data}
\end{table}

\paragraph{Evaluation Metric} The measure $\text{R@}n,\text{IoU@}\mu$ denotes the percentage of test samples that have at least one result whose Intersection over Union (IoU) with ground truth is larger than $\mu$ in top-$n$ predictions. The measure  $\text{dR@}n,\text{IoU@}\mu$, is the discounted-$\text{R@}n,\text{IoU@}\mu$ proposed by~\citet{yuan2021closer}. This metric introduces two discount factors computed from the boundaries of predicted and ground truth moments to restrain over-long predictions. We also report mIoU, which is the average IoU over all test samples. We set $n=1$ and $\mu\in\{0.3, 0.5, 0.7\}$. 

\paragraph{Implementation Details} We utilize the $300$d GloVe~\cite{pennington2014glove} vectors to initialize the words in query $Q$. For the video $V$, we follow~\citet{yuan2021closer} with pre-trained I3D features~\cite{Carreira2017QuoVA} for Cha-CD and Cha-STA, and C3D features~\cite{tran2015learning} for ANet-CD and ANetCap. For the model parameters, we follow~\citet{zhang2020vslnet} and use the hidden size of $128$ for all hidden layers, the head size of 8 for multi-head attention, and the kernel size of 7 for convolutions. For data debiasing, we empirically partition each video into $N_{clip}=5$ clips. Adam~\cite{Kingma2015AdamAM} optimizer is used with batch size of $16$ and learning rate of $0.0005$ for training.

\subsection{Experimental Results}

\begin{table}[t]
    \small
    \centering
	\setlength{\tabcolsep}{4.2 pt}
	\begin{tabular}{c | c | c c | c c | c}
		\specialrule{.1em}{.05em}{.05em}
		\multirow{2}{*}{Split} & \multirow{2}{*}{Method} & \multicolumn{2}{c |}{$\text{R@}1,\text{IoU@}\mu$} & \multicolumn{2}{c |}{$\text{dR@}1,\text{IoU@}\mu$} & \multirow{2}{*}{mIoU} \\
        & & $\mu$=0.5 & $\mu$=0.7 & $\mu$=0.5 & $\mu$=0.7 & \\
        \specialrule{.1em}{.05em}{.05em}
        \multirow{3}{*}{iid} & Q-only & 29.65 & 17.89 & 26.98 & 17.06 & 34.67 \\
                             & V-only & 30.38 & 20.29 & 27.94 & 19.28 & 33.24 \\
                             & VSLNet & 60.51 & 41.07 & 56.12 & 39.25 & 54.39 \\
        \hline
        \multirow{3}{*}{ood} & Q-only & 15.17 &  5.99 & 12.96 & 5.54 & 20.87 \\
                             & V-only & 16.71 &  6.96 & 14.18 & 6.48 & 20.03 \\
                             & VSLNet & 48.18 & 28.89 & 43.29 & 27.20 & 45.56 \\
        \specialrule{.1em}{.05em}{.05em}
	\end{tabular}
	\caption{\small The performance (\%) of unimodal models and VSLNet on Charades-CD dataset.}
	\label{tab:charades_cd_bias_exp}
\end{table}

\paragraph{Bias in backbone model}
We first study the performance of backbone model on the Cha-CD dataset. In particular, we separately train the standard VSLNet, query-only model (Q-only), and video-only model (V-only). The results are summarized in Table~\ref{tab:charades_cd_bias_exp}. Observe that both Q-only and V-only models achieve fair performance on iid test set, but poorer on ood test set, \eg $17.06\%$ on iid versus $5.54\%$ on ood for Q-only model over the $\text{dR@}n,\text{IoU@}0.7$ measure. Similar observation holds on the standard VSLNet. We conclude that the backbone model fails to generalize well on test set with ood samples.

\begin{table}[t]
    \small
    \centering
	\setlength{\tabcolsep}{1 pt}
	\begin{tabular}{c | c | c c | c | c c | c}
		\specialrule{.1em}{.05em}{.05em}
		\multirow{3}{*}{Split} & \multirow{3}{*}{Method} & \multicolumn{3}{c |}{Charades-CD} & \multicolumn{3}{c}{ActivityNet-CD} \\
		\cline{3-8}
		& & \multicolumn{2}{c |}{$\text{dR@}1,\text{IoU@}\mu$} & \multirow{2}{*}{mIoU} & \multicolumn{2}{c |}{$\text{dR@}1,\text{IoU@}\mu$} & \multirow{2}{*}{mIoU} \\
        & & $\mu$=0.5 & $\mu$=0.7 & & $\mu$=0.5 & $\mu$=0.7 & \\
        \specialrule{.1em}{.05em}{.05em}
        \multirow{4}{*}{iid} & VSLNet & \textbf{56.12} & \textbf{39.25} & \textbf{54.39} & \textbf{43.40} & \textbf{31.37} & \textbf{47.57} \\
                             & +DD    & 55.01 & 36.78 & 53.42 & \textit{42.45} & \textit{28.28} & \textit{45.90} \\
                             & +MD    & 54.69 & 37.41 & 53.26 & 40.51 & 27.60 & 43.62 \\
                             & +DD+MD & \textit{55.66} & \textit{38.87} & \textit{53.92} & 40.88 & 28.11 & 44.71 \\
        \hline
        \multirow{4}{*}{ood} & VSLNet & 43.29 & 27.20 & 45.56 & 20.96 & 11.59 & 29.95 \\
                             & +DD    & 49.02 & \textit{31.69} & \textit{49.60} & 23.09 & 13.45 & 30.19 \\
                             & +MD    & \textit{49.26} & 31.64 & 49.49 & \textbf{25.33} & \textit{14.41} & \textit{30.55} \\
                             & +DD+MD & \textbf{50.37} & \textbf{32.70} & \textbf{50.30} & \textit{25.05} & \textbf{14.67} & \textbf{30.56} \\
        \hline
        \multirow{4}{*}{all} & VSLNet & 45.81 & 29.56 & 47.30 & 25.47 & 15.66 & 33.19 \\
                             & +DD    & 50.19 & 32.81 & \textit{50.57} & 26.77 & 16.43 & 33.31 \\
                             & +MD    & \textit{50.63} & \textit{33.33} & 50.40 & \textbf{28.38} & \textit{17.05} & \textit{33.17} \\
                             & +DD+MD & \textbf{51.40} & \textbf{34.08} & \textbf{51.01} & \textit{28.25} & \textbf{17.43} & \textbf{33.42} \\
        \specialrule{.1em}{.05em}{.05em}
	\end{tabular}
	\caption{\small The performance (\%) of data debiasing (DD) and model debiasing (MD) strategies on Charades-CD and ActivityNet-CD datasets.}
	\label{tab:dd_md_results_cd}
\end{table}

\begin{table}[t]
    \small
    \centering
	\setlength{\tabcolsep}{2.8 pt}
	\begin{tabular}{c | c c | c | c c | c}
		\specialrule{.1em}{.05em}{.05em}
		\multirow{3}{*}{Method} & \multicolumn{3}{c |}{Charades-STA} & \multicolumn{3}{c}{ActivityNet Caps} \\
		\cline{2-7}
		& \multicolumn{2}{c |}{$\text{dR@}1,\text{IoU@}\mu$} & \multirow{2}{*}{mIoU} & \multicolumn{2}{c |}{$\text{dR@}1,\text{IoU@}\mu$} & \multirow{2}{*}{mIoU} \\
        & $\mu$=0.5 & $\mu$=0.7 & & $\mu$=0.5 & $\mu$=0.7 & \\
        \specialrule{.1em}{.05em}{.05em}
        VSLNet  & 51.22 & 35.08 & 51.31 & \textit{37.16} & 24.29 & 43.01 \\
        +DD     & 51.12 & \textit{37.10} & \textit{51.96} & 36.88 & 24.45 & \textbf{43.31} \\
        +MD     & \textit{51.55} & 36.05 & 51.17 & \textit{37.16} & \textit{25.03} & 42.88 \\
        +DD+MD & \textbf{53.97} & \textbf{38.06} & \textbf{52.18} & \textbf{37.45} & \textbf{25.26} & \textit{43.26} \\
        \specialrule{.1em}{.05em}{.05em}
	\end{tabular}
	\caption{\small Results (in \%) of data debiasing (DD) and model debiasing (MD) strategies on Charades-STA and ActivityNet Captions datasets.}
	\label{tab:dd_md_results_org}
\end{table}

\paragraph{Impact of data and model debiasing}
Results of applying data debiasing (DD) and model debiasing (MD) strategies are summarized in Table~\ref{tab:dd_md_results_cd}, where ``all'' means all samples in both iid and ood test sets.\footnote{More complete results are summarized in Appendix.} On both iid test sets, performance drop is observed after applying DD and MD strategies. 
The reason is both DD and MD prevent the model from exploiting distribution bias in training. On contrast, for ood test sets, both DD and MD significantly improve the performance. In addition, by combining DD and MD, further improvements are observed on ood test sets. Thus, the performance gaps between iid and ood sets are reduced further. On the combined test samples (iid + ood), DD and MD together bring performance increase on both datasets.  

Table~\ref{tab:dd_md_results_org} reports results on original Cha-STA and ANetCap datasets. Note moment annotation distributions of train and test sets are almost the same for both Cha-STA and ANetCap datasets~\cite{yuan2021closer}. Observe that small improvements are observed after applying DD and MD strategies on both datasets. The DD and MD encourage TSGV model to focus more on exploiting cross-modal reasoning between video and query, which contributes to the performance improvements.

\begin{figure*}[t]
    \centering
	\includegraphics[trim={0cm 0.4cm 0cm 0.0cm},clip,width=\textwidth]{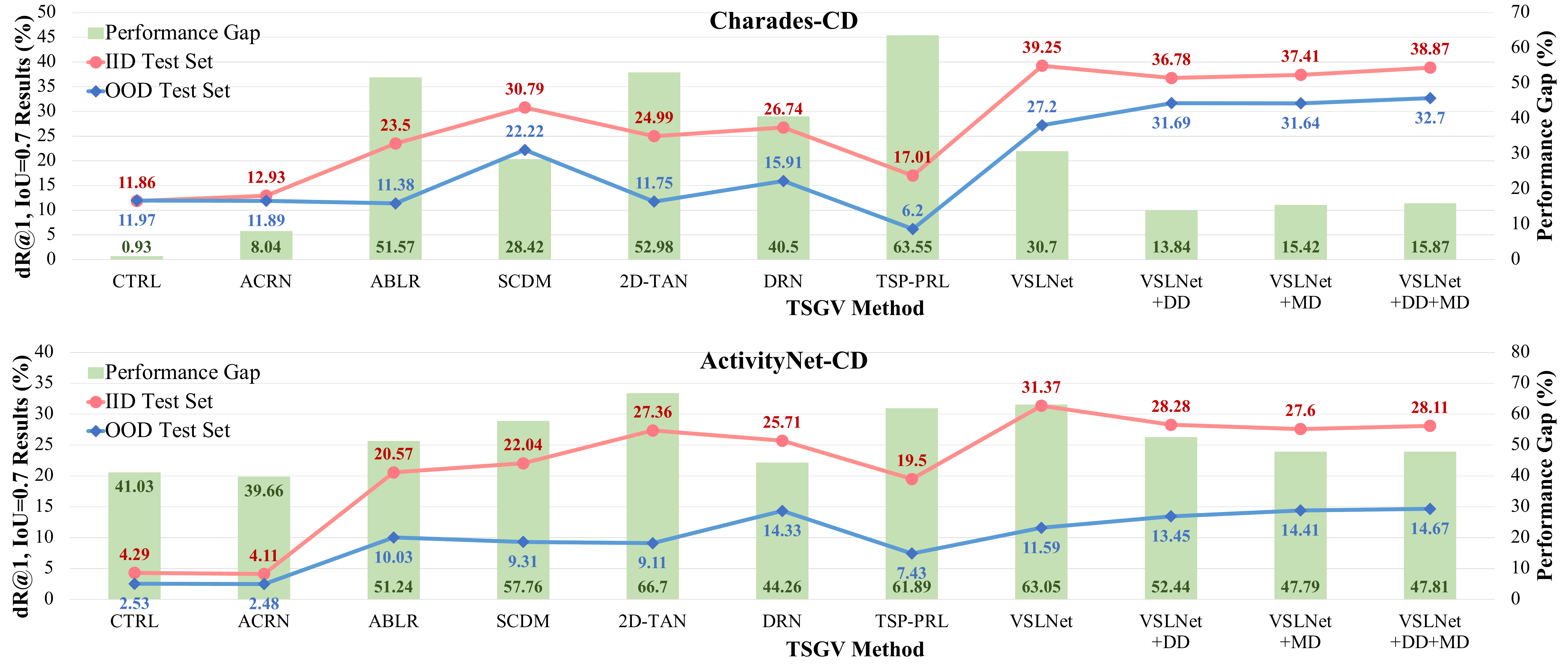}
	\caption{\small Results (in \%) of $\text{dR@}1,\text{IoU@}0.7$ and the performance gap (\%) between iid and ood test sets for SOTA TSGV models (the results are reported by~\citet{yuan2021closer}), VSLNet and the proposed debiasing strategies on Charades-CD and ActivityNet-CD datasets.}
	\label{fig:sota_compare}
\end{figure*}

\begin{table}[t]
    \small
    \centering
	\begin{tabular}{c | c | c c c | c}
		\specialrule{.1em}{.05em}{.05em}
		\multirow{2}{*}{Split} & \multirow{2}{*}{$N_{clip}$} & \multicolumn{3}{c |}{$\text{dR@}1,\text{IoU@}\mu$} & \multirow{2}{*}{mIoU} \\
        & & $\mu$=0.3 & $\mu$=0.5 & $\mu$=0.7 & \\
        \specialrule{.1em}{.05em}{.05em}
        \multirow{3}{*}{iid} & 4 & \textbf{65.82} & 54.75 & \textbf{37.46} & 53.75 \\
                             & 5 & \textit{65.79} & \textit{55.01} & 36.78 & \textbf{53.92} \\
                             & 6 & 65.67 & \textbf{56.46} & \textit{36.86} & \textit{53.86} \\
        \hline
        \multirow{3}{*}{ood} & 4 & \textit{59.89} & 48.63 & \textit{31.31} & 48.86 \\
                             & 5 & \textbf{60.68} & \textbf{49.02} & \textbf{31.69} & \textbf{49.60} \\
                             & 6 & 59.45 & \textit{48.81} & 30.80 & \textit{49.33} \\
        \hline
        \multirow{3}{*}{all} & 4 & \textit{61.05} & 50.09 & \textit{32.51} & 49.82 \\
                             & 5 & \textbf{61.68} & \textbf{50.19} & \textbf{32.81} & \textbf{50.57} \\
                             & 6 & 60.67 & \textit{50.31} & 31.99 & \textit{50.22} \\
        \specialrule{.1em}{.05em}{.05em}
	\end{tabular}
	\caption{\small The performance (\%) of VSLNet with data debiasing over different number of clip $N_{clip}$ on Charades-CD dataset.}
	\label{tab:n_clip_analysis}
\end{table}

\begin{table}[t]
    \small
    \centering
	\begin{tabular}{c | c | c c c | c}
		\specialrule{.1em}{.05em}{.05em}
		\multirow{2}{*}{Split} & \multirow{2}{*}{Method} & \multicolumn{3}{c |}{$\text{dR@}1,\text{IoU@}\mu$} & \multirow{2}{*}{mIoU} \\
        & & $\mu$=0.3 & $\mu$=0.5 & $\mu$=0.7 & \\
        \specialrule{.1em}{.05em}{.05em}
        \multirow{3}{*}{iid} & VSLNet & \textbf{66.72} & \textit{56.12} & \textbf{39.25} & \textbf{54.39} \\
                             & +V-MD  & \textit{65.68} & \textbf{56.41} & 38.04 & \textit{54.11} \\
                             & +Q-MD  & 65.56 & 55.95 & \textit{38.95} & 53.41 \\
                             & +MD    & 65.22 & 54.69 & 37.41 & 53.26 \\
        \hline
        \multirow{3}{*}{ood} & VSLNet & 55.85 & 43.29 & 27.20 & 45.56 \\
                             & +V-MD  & \textit{58.21} & \textit{47.50} & \textit{30.53} & \textit{48.42} \\
                             & +Q-MD  & 57.90 & 46.63 & 30.33 & 47.60 \\
                             & +MD    & \textbf{60.59} & \textbf{49.26} & \textbf{31.64} & \textbf{49.49} \\
        \hline
        \multirow{3}{*}{all} & VSLNet & 57.98 & 45.81 & 29.56 & 47.30 \\
                             & +V-MD  & \textit{59.67} & \textit{49.25} & 32.00 & \textit{49.53} \\
                             & +Q-MD  & 59.40 & 48.45 & \textit{32.02} & 48.74 \\
                             & +MD    & \textbf{60.93} & \textbf{50.63} & \textbf{33.33} & \textbf{50.40} \\
        \specialrule{.1em}{.05em}{.05em}
	\end{tabular}
	\caption{\small The performance (\%) of VSLNet with different model debiasing strategies on Charades-CD dataset.}
	\label{tab:md_analysis}
\end{table}

\paragraph{Comparison with state-of-the-arts}
Figure~\ref{fig:sota_compare} depicts the $\text{dR@}1,\text{IoU@}0.7$ results of SOTA models on Cha-CD and ANet-CD datasets. We compare with 1) the proposal-based methods, CTRL~\cite{Gao2017TALLTA}, ACRN~\cite{Liu2018AMR} and SCDM~\cite{yuan2019semantic}; 2) the proposal-free methods, ABLR~\cite{Yuan2019ToFW}, 2D-TAN~\cite{zhang2020learning} and DRN~\cite{zeng2020dense}; 3) the RL-based method TSP-PRL~\cite{Wu2020TreeStructuredPB}. All the results of these models are reported by~\citet{yuan2021closer}. In general, our backbone model, and the version with DD and MD strategies, are superior to the compared SOTA models on both datasets. 

Although the performance of these models varies greatly, we are more interested in their performance gap between iid and ood test sets. Specifically, we define the performance gap as:
\begin{equation}
    p_{gap}=\frac{|s_{ood}-s_{iid}|}{s_{iid}}\times 100\%
\nonumber
\end{equation}
where $|\cdot|$ denotes absolute value operation, $s_{iid}$ and $s_{ood}$ represent the score of a model on iid and ood test sets, respectively. Smaller $p_{gap}$ means the model is more robust.

On Cha-CD dataset, CTRL and ACRN achieve comparable performance gap between iid and ood test sets. Other methods, including VSLNet, show large performance gap between iid and ood test sets. Compared to those SOTAs, the gap of VSLNet is moderate, with $p_{gap}=30.70\%$. With data debiasing, $p_{gap}$ of VSLNet decreases from $30.70\%$ to $13.84\%$. DD strategy significantly improves the results on ood test set by balancing the distribution of train set to be more uniform. MD strategy also reduces the $p_{gap}$ distinctly by disentangling the bias from the TSGV model with two unimodal branches during training.

On ANet-CD dataset, the performance gap between iid and ood test sets are more conspicuous than that on Cha-CD dataset. Specifically, $p_{gap}$ of VSLNet is $63.05\%$, the second largest among all models. By applying DD and MD strategies, the results on ood set increase and the results on iid set slightly decrease. After debiasing, $p_{gap}$ of VSLNet reduces significantly.

\paragraph{Analysis of data debiasing strategy}
We now study the effect of the number of clips $N_{clip}$ in data debiasing (DD) strategy on Cha-CD dataset. We evaluate $N_{clip}\in\{4,5,6\}$ and report results in Table~\ref{tab:n_clip_analysis}. Observe that performances of VSLNet+DD with different $N_{clip}$s are comparable on the iid test set. However, VSLNet+DD with $N_{clip}=5$ consistently outperforms the model with other $N_{clip}$ values on the ood test set. Besides, VSLNet+DD with $N_{clip}=5$ also performs best over all test samples. Similar observations hold on the ANet-CD dataset.

\paragraph{Analysis of model debiasing strategy}
Here we study the effect of each unimodal model, \ie video-only and query-only branches on VSLNet. The results are summarized in Table~\ref{tab:md_analysis}. V-MD and Q-MD denotes debiasing using video-only or query-only branch only. Both V-MD and Q-MD strategies lead to significant improvement on the ood test set, and slight degradation on the iid test set. We observe the same on the ANet-CD dataset. This set of results indicate that model debiasing on both video and query sides are necessary for learning a robust TSGV model.

\begin{table}[t]
    \small
    \centering
    \setlength{\tabcolsep}{3 pt}
	\begin{tabular}{c | c c c c c c}
		\specialrule{.1em}{.05em}{.05em}
		Dataset & $\bar{L}_{V}$ & $\bar{L}_{M}$ & $\bar{L}_{Q}$ & $\bar{N}_{A/V}$ & $N_{vocab}$ & $N_{act}$ \\
        \specialrule{.1em}{.05em}{.05em}
        Cha-CD  &  30.75s &  8.12s &  6.22 & 2.42 &  1,255 & 69 \\
        ANet-CD & 117.60s & 37.14s & 14.41 & 4.82 & 13,707 & 901 \\
        \specialrule{.1em}{.05em}{.05em}
	\end{tabular}
	\caption{\small Data statistics of Charades-CD and ActivityNet-CD. $\bar{L}_{V}$/ $\bar{L}_{M}$ is the average video/moment length in seconds, $\bar{L}_{Q}$ is average number of words in query, $\bar{N}_{A/V}$ is average annotations per video, $N_{vocab}$ is the vocabulary size and $N_{act}$ is the size action verb. Note we only count the verb with occurrence larger than 5 for $N_{act}$.}
	\label{tab:data_stat}
\end{table}

\begin{figure}[t]
    \centering
	\includegraphics[width=0.47\textwidth]{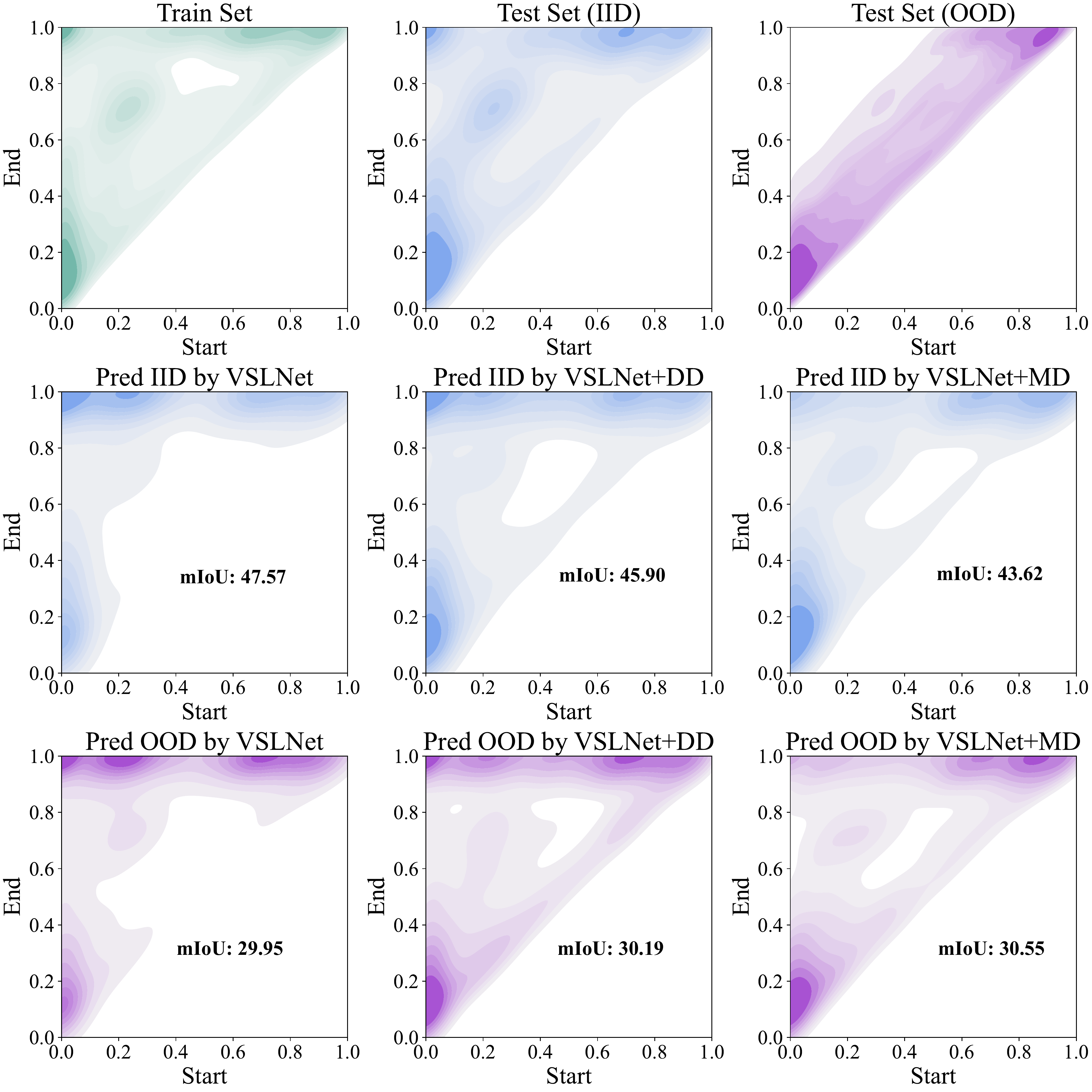}
	\caption{\small Visualization of moment annotation distributions of train, iid and ood test sets in ActivityNet-CD dataset, and that predicted by VSLNet and the proposed debiasing strategies.}
	\label{fig:visual_anet}
\end{figure}

\paragraph{Performance differences on Cha-CD and ANet-CD}
The data and model debiasing strategies show their strong generalization capability on the test set with out-of-distribution. However, we observe that the performance of the two strategies on ANet-CD dataset are inferior to that on Cha-CD dataset. Based on data statistics of the two datasets (see Table~\ref{tab:data_stat}), ANet-CD dataset is more challenging than Cha-CD, because the average video and query lengths of ANet-CD are much larger. Besides, ANet-CD contains more than 900 different action verbs while Cha-CD has 69 verbs only. That is, the activities/events of ANet-CD are many more diverse than that in Cha-CD. This could contribute to the difficulty of debiasing ANet-CD. Figure~\ref{fig:visual_anet} depicts the predicted annotation distributions of VSLNet and two proposed debiasing strategies on iid and ood test sets. Despite the improvements have been made by DD and MD, we observe that the predicted distributions of ood test set by VSLNet+DD/+MD remain similar to that of train set. In other words, the bias issue is more challenging to address on ANet-CD dataset. 

\section{Conclusion}\label{sec:conclusion}
In this work, we propose two simple yet effective strategies, data debiasing and model debiasing, to mitigate data distribution bias  in TSGV task. We design the data debiasing strategy to balance the sample distribution by oversampling with video truncation. The  model debiasing guides the TSGV model to learn accurate cross-modal interactions. With the help of two  unimodal branches, we reduce the loss propagated to the TSGV model for biased samples and amplify the loss for non-biased samples. Through extensive experiments, we show that both data and model debiasing strategies contribute performance improvement on ood test sets. 


\bibliography{emnlp}
\bibliographystyle{acl_natbib}

\appendix
\section{Detailed Experimental Settings and Results}
\label{sec:appendix}
This appendix provides detailed experiment settings, the visualization of the benchmark datasets and more complete results of different strategies on the benchmark datasets.

We utilize the $300$d GloVe~\cite{pennington2014glove} vectors to initialize the lowercase words in query $Q$. For the video $V$, we follow~\citet{yuan2021closer} with pre-trained I3D features~\cite{Carreira2017QuoVA} for Cha-CD and Cha-STA, and pre-trained C3D features~\cite{tran2015learning} for ANet-CD and ANetCap. For all the benchmark datasets, we set the maximal visual sequence feature length as $128$. For the model parameters, we follow~\citet{zhang2020vslnet} and use the hidden size of $128$ for all hidden layers, the head size of 8 for multi-head attention, and the kernel size of 7 for convolutions. For data debiasing, we empirically partition each video into $N_{clip}=5$ clips. Adam~\cite{Kingma2015AdamAM} optimizer is used with batch size of $16$ and learning rate of $0.0005$ for training. The model is trained by 100 epochs in total with early stopping tolerance of $10$ epochs. All experiments are conducted on a workstation with dual NVIDIA GeForce RTX 3090 GPUs.

\begin{figure*}[htbp!]
    \centering
    \subfigure[\small Charades-STA.]
	{\label{fig:full_charades_sta_dist}	\includegraphics[width=0.38\textwidth]{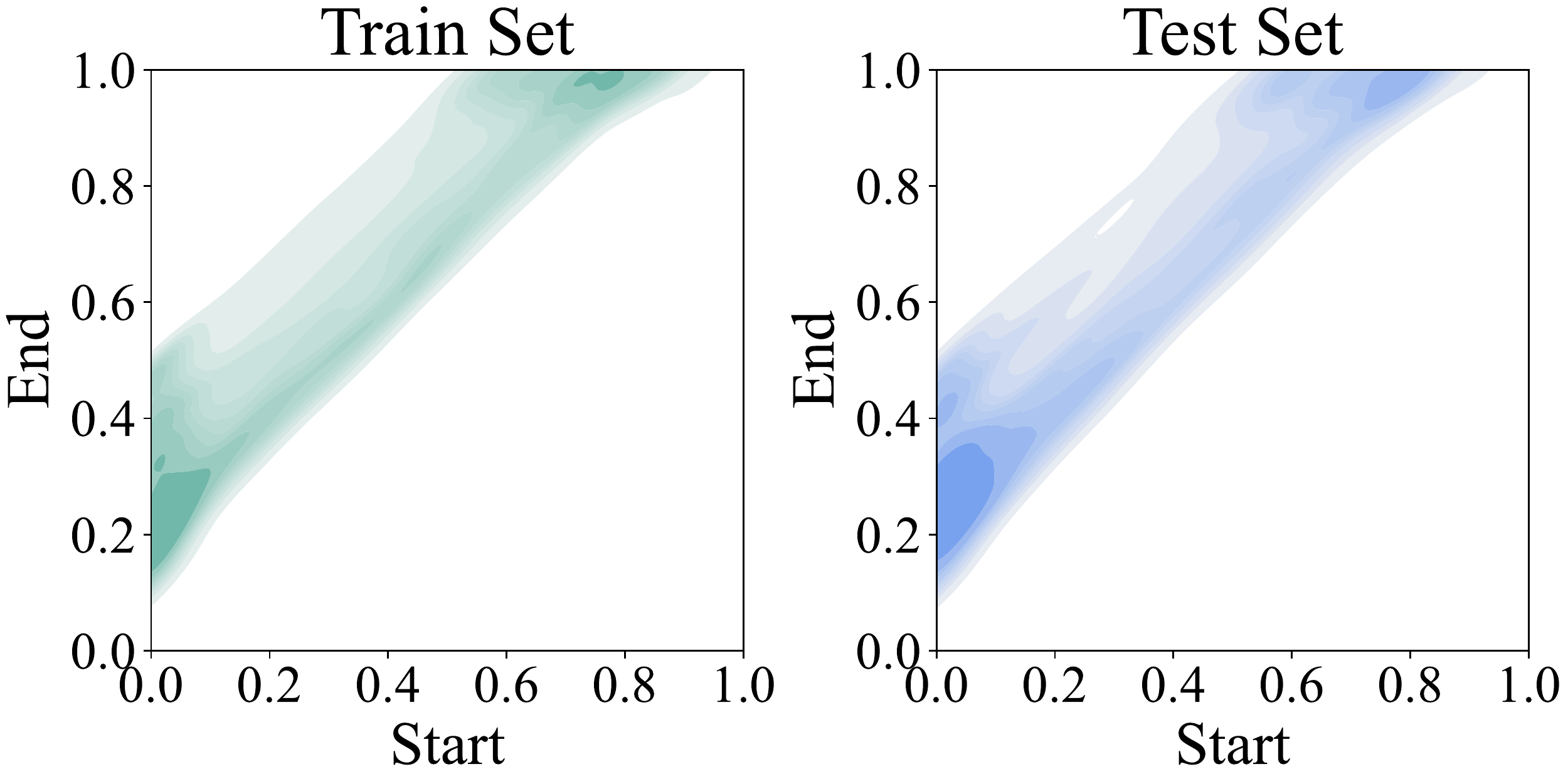}}
	\subfigure[\small Charades-CD.]
	{\label{fig:full_charades_cd_dist}	\includegraphics[width=0.57\textwidth]{figures/charades_cd_dist}}
	\subfigure[\small ActivityNet Captions.]
	{\label{fig:full_activitynet_org_dist}	\includegraphics[width=0.38\textwidth]{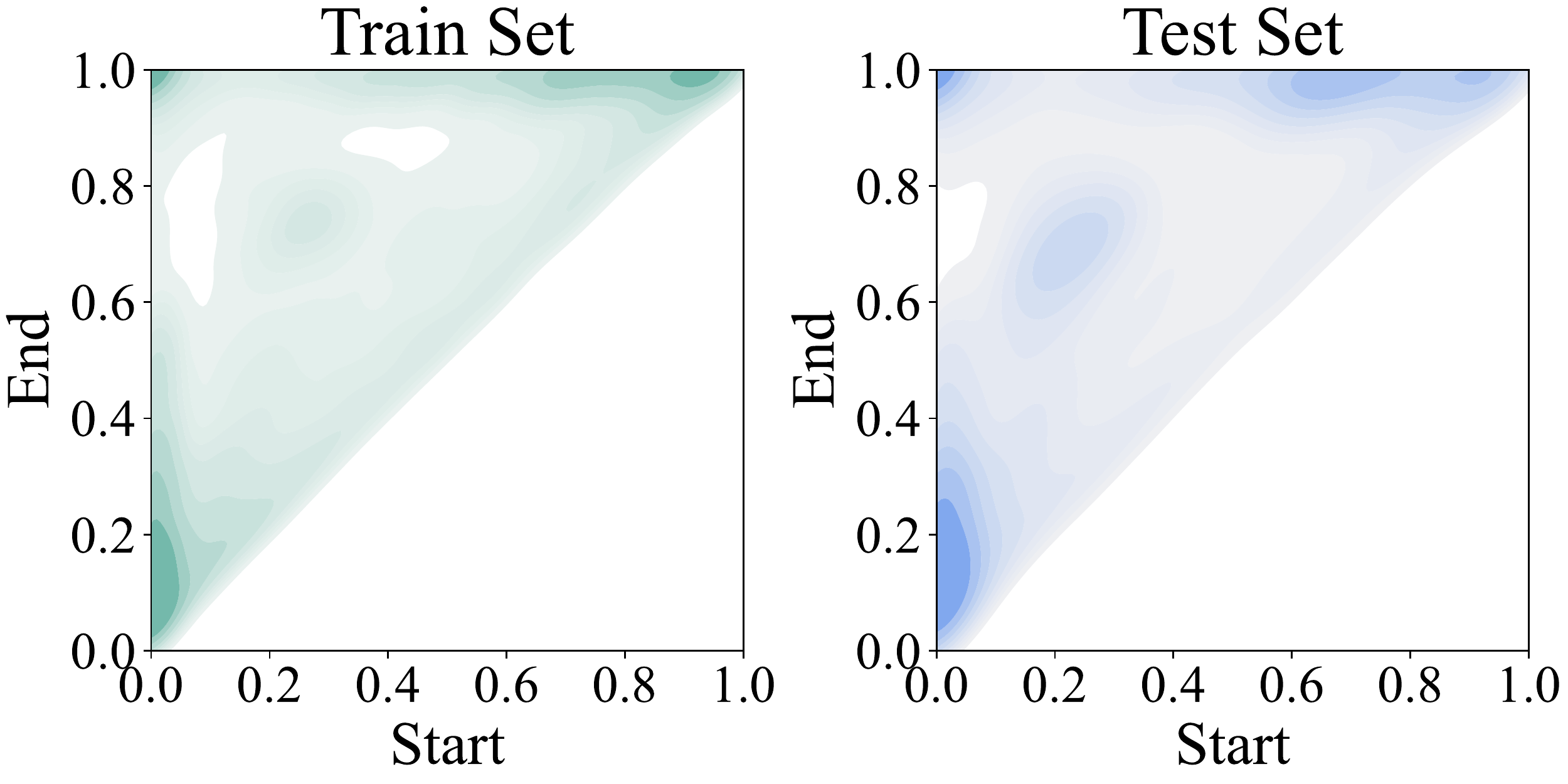}}
	\subfigure[\small ActivityNet-CD]
	{\label{fig:full_activitynet_cd_dist}	\includegraphics[width=0.57\textwidth]{figures/activitynet_cd_dist}}
	\caption{\small Moment annotation distributions of Charades-STA~\cite{Gao2017TALLTA}, Charades-CD~\cite{yuan2021closer}, ActivityNet Captions~\cite{krishna2017dense} and ActivityNet-CD~\cite{yuan2021closer} datasets, where ``Start'' and ``End'' axes represent the normalized start and end time points, respectively. The deeper the color, the larger density (\ie more annotations) in the dataset.}
	\label{fig:full_cd_dist}
\end{figure*}

Figure~\ref{fig:full_cd_dist} depicts the moment annotation distributions of the evaluated datasets, \ie Charades-STA~\cite{Gao2017TALLTA}, Charades-CD~\cite{yuan2021closer}, ActivityNet Captions~\cite{krishna2017dense} and ActivityNet-CD~\cite{yuan2021closer}. 
Table~\ref{tab:full_dd_md_results_cd} summarizes the complete results of $\text{R@}n, \text{IoU@}\mu$, $\text{dR@}n, \text{IoU@}\mu$ and mIoU by applying data debiasing and model debiasing strategies on top of VSLNet on the Charades-CD and ActivityNet-CD datasets. Table~\ref{tab:full_dd_md_results_org} summarizes the complete results of $\text{R@}n, \text{IoU@}\mu$, $\text{dR@}n, \text{IoU@}\mu$ and mIoU by applying data debiasing and model debiasing strategies on top of VSLNet on the Charades-STA and ActivityNet Captions datasets.
Table~\ref{tab:full_n_clip_analysis} summarizes the complete results of VSLNet with data debiasing strategy over different number of clip $N_{clip}$ on the Charades-CD and ActivityNet-CD datasets. Table~\ref{tab:full_md_analysis} summarizes the complete results of VSLNet with different model debiasing strategies on the Charades-CD and ActivityNet-CD datasets.

\begin{table*}[htbp!]
    \small
    \centering
	\setlength{\tabcolsep}{3 pt}
	\begin{tabular}{c | c | c c c | c c c | c | c c c | c c c | c}
		\specialrule{.1em}{.05em}{.05em}
		\multirow{3}{*}{Split} & \multirow{3}{*}{Method} & \multicolumn{7}{c |}{Charades-CD} & \multicolumn{7}{c}{ActivityNet-CD} \\
		\cline{3-16}
		& & \multicolumn{3}{c |}{$\text{R@}1,\text{IoU@}\mu$} & \multicolumn{3}{c |}{$\text{dR@}1,\text{IoU@}\mu$} & \multirow{2}{*}{mIoU} & \multicolumn{3}{c |}{$\text{dR@}1,\text{IoU@}\mu$} & \multicolumn{3}{c |}{$\text{dR@}1,\text{IoU@}\mu$} & \multirow{2}{*}{mIoU} \\
        & & $\mu$=0.3 & $\mu$=0.5 & $\mu$=0.7 & $\mu$=0.3 & $\mu$=0.5 & $\mu$=0.7 & & $\mu$=0.3 & $\mu$=0.5 & $\mu$=0.7 & $\mu$=0.3 & $\mu$=0.5 & $\mu$=0.7 & \\
        \specialrule{.1em}{.05em}{.05em}
        \multirow{4}{*}{iid} & VSLNet & 75.09 & 60.51 & 41.07 & 66.72 & 56.12 & 39.25 & 54.39 & 63.18 & 49.16 & 32.37 & 52.41 & 43.40 & 31.37 & 47.57 \\
                             & +DD    & 74.73 & 59.54 & 38.40 & 65.79 & 55.01 & 36.78 & 53.42 & 62.08 & 47.95 & 30.10 & 51.81 & 42.45 & 28.28 & 45.90 \\
                             & +MD    & 73.39 & 59.17 & 39.13 & 65.22 & 54.69 & 37.41 & 53.26 & 58.73 & 45.10 & 29.32 & 50.23 & 40.51 & 27.60 & 43.62 \\
                             & +DD+MD & 74.85 & 59.97 & 40.55 & 67.19 & 55.66 & 38.87 & 53.92 & 60.65 & 45.84 & 29.94 & 51.26 & 40.88 & 28.11 & 44.71 \\
        \hline
        \multirow{4}{*}{ood} & VSLNet & 65.93 & 48.18 & 28.89 & 55.85 & 43.29 & 27.20 & 45.56 & 41.71 & 23.31 & 12.06 & 32.99 & 20.96 & 11.59 & 29.95 \\
                             & +DD    & 71.88 & 54.84 & 33.69 & 60.68 & 49.02 & 31.69 & 49.60 & 42.21 & 25.09 & 14.00 & 35.12 & 23.09 & 13.45 & 30.19 \\
                             & +MD    & 70.79 & 54.90 & 33.30 & 60.59 & 49.26 & 31.64 & 49.49 & 42.43 & 27.70 & 15.01 & 36.08 & 25.33 & 14.41 & 30.55 \\
                             & +DD+MD & 70.10 & 55.82 & 34.70 & 60.81 & 50.37 & 32.70 & 50.30 & 42.78 & 27.33 & 15.10 & 36.16 & 25.05 & 14.67 & 30.56 \\
        \hline
        \multirow{4}{*}{all} & VSLNet & 67.72 & 50.60 & 31.28 & 57.98 & 45.81 & 29.56 & 47.30 & 46.02 & 28.49 & 16.14 & 36.89 & 25.47 & 15.66 & 33.19 \\
                             & +DD    & 72.44 & 55.76 & 34.61 & 61.68 & 50.19 & 32.81 & 50.57 & 46.19 & 29.68 & 17.23 & 38.18 & 26.77 & 16.43 & 33.31 \\
                             & +MD    & 70.25 & 55.79 & 35.30 & 60.93 & 50.63 & 33.33 & 50.40 & 45.70 & 31.19 & 17.88 & 38.92 & 28.38 & 17.05 & 33.17 \\
                             & +DD+MD & 71.03 & 57.03 & 36.06 & 62.06 & 51.40 & 34.08 & 51.01 & 46.38 & 31.09 & 18.14 & 39.21 & 28.25 & 17.43 & 33.42 \\
        \specialrule{.1em}{.05em}{.05em}
	\end{tabular}
	\caption{\small The performance (\%) of applying data debiasing (DD) and model debiasing (MD) strategies on top of VSLNet on the Charades-CD and ActivityNet-CD datasets.}
	\label{tab:full_dd_md_results_cd}
\end{table*}

\begin{table*}[htbp!]
    \small
    \centering
	\setlength{\tabcolsep}{3.6 pt}
	\begin{tabular}{c | c c c | c c c | c | c c c | c c c | c}
		\specialrule{.1em}{.05em}{.05em}
		\multirow{3}{*}{Method} & \multicolumn{7}{c |}{Charades-STA} & \multicolumn{7}{c}{ActivityNet Captions} \\
		\cline{2-15}
		& \multicolumn{3}{c |}{$\text{R@}1,\text{IoU@}\mu$} & \multicolumn{3}{c |}{$\text{dR@}1,\text{IoU@}\mu$} & \multirow{2}{*}{mIoU} & \multicolumn{3}{c |}{$\text{dR@}1,\text{IoU@}\mu$} & \multicolumn{3}{c |}{$\text{dR@}1,\text{IoU@}\mu$} & \multirow{2}{*}{mIoU} \\
        & $\mu$=0.3 & $\mu$=0.5 & $\mu$=0.7 & $\mu$=0.3 & $\mu$=0.5 & $\mu$=0.7 & & $\mu$=0.3 & $\mu$=0.5 & $\mu$=0.7 & $\mu$=0.3 & $\mu$=0.5 & $\mu$=0.7 & \\
        \specialrule{.1em}{.05em}{.05em}
        VSLNet & 71.18 & 55.73 & 36.83 & 62.04 & 51.22 & 35.08 & 51.31 & 59.30 & 42.50 & 26.01 & 48.13 & 37.16 & 24.29 & 43.01 \\
        +DD    & 71.75 & 55.38 & 38.90 & 62.49 & 51.12 & 37.10 & 51.96 & 59.68 & 42.21 & 26.19 & 48.18 & 36.88 & 24.45 & 43.31 \\
        +MD    & 70.19 & 55.75 & 37.80 & 61.72 & 51.55 & 36.05 & 51.17 & 58.46 &  42.14 & 26.74 & 47.96 & 37.16 & 25.03 & 42.88 \\
        +DD+MD & 70.05 & 58.49 & 39.92 & 62.27 & 53.97 & 38.06 & 52.18 & 59.66 & 42.89 & 26.99 & 48.99 & 37.45 & 25.26 & 43.46 \\
        \specialrule{.1em}{.05em}{.05em}
	\end{tabular}
	\caption{\small The performance (\%) of applying data debiasing (DD) and model debiasing (MD) strategies on top of VSLNet on the Charades-STA and ActivityNet Captions datasets.}
	\label{tab:full_dd_md_results_org}
\end{table*}

\begin{table*}[htbp!]
    \small
    \centering
    \setlength{\tabcolsep}{3.4 pt}
	\begin{tabular}{c | c | c c c | c c c | c | c c c | c c c | c}
		\specialrule{.1em}{.05em}{.05em}
		\multirow{3}{*}{Split} & \multirow{3}{*}{$N_{clip}$} & \multicolumn{7}{c |}{Charades-CD} & \multicolumn{7}{c}{ActivityNet-CD} \\
		\cline{3-16}
		& & \multicolumn{3}{c |}{$\text{R@}1,\text{IoU@}\mu$} & \multicolumn{3}{c |}{$\text{dR@}1,\text{IoU@}\mu$} & \multirow{2}{*}{mIoU} & \multicolumn{3}{c |}{$\text{R@}1,\text{IoU@}\mu$} & \multicolumn{3}{c |}{$\text{dR@}1,\text{IoU@}\mu$} & \multirow{2}{*}{mIoU} \\
        & & $\mu$=0.3 & $\mu$=0.5 & $\mu$=0.7 & $\mu$=0.3 & $\mu$=0.5 & $\mu$=0.7 & & $\mu$=0.3 & $\mu$=0.5 & $\mu$=0.7 & $\mu$=0.3 & $\mu$=0.5 & $\mu$=0.7 & \\
        \specialrule{.1em}{.05em}{.05em}
        \multirow{3}{*}{iid} & 4 & 74.00 & 59.17 & 39.13 & 65.82 & 54.75 & 37.46 & 53.75 & 61.49 & 46.49 & 28.90 & 51.36 & 41.15 & 27.16 & 45.17 \\
                             & 5 & 74.73 & 59.54 & 38.40 & 65.79 & 55.01 & 36.78 & 53.92 & 63.18 & 47.11 & 30.81 & 52.23 & 41.63 & 28.88 & 45.98 \\
                             & 6 & 74.24 & 61.24 & 38.52 & 65.67 & 56.46 & 36.86 & 53.86 & 61.92 & 46.30 & 30.45 & 51.46 & 41.21 & 28.61 & 45.63 \\
        \hline
        \multirow{3}{*}{ood} & 4 & 70.43 & 54.28 & 33.24 & 59.89 & 48.63 & 31.31 & 48.86 & 42.56 & 26.47 & 14.07 & 35.31 & 24.04 & 13.45 & 30.39 \\
                             & 5 & 71.88 & 54.84 & 33.69 & 60.68 & 49.02 & 31.69 & 49.60 & 43.09 & 26.81 & 14.33 & 35.63 & 24.29 & 13.74 & 31.30 \\
                             & 6 & 69.66 & 54.52 & 32.71 & 59.45 & 48.81 & 30.80 & 49.33 & 42.62 & 26.04 & 13.68 & 35.08 & 23.73 & 13.16 & 30.33 \\
        \hline
        \multirow{3}{*}{all} & 4 & 71.13 & 55.55 & 34.40 & 61.05 & 50.09 & 32.51 & 49.82 & 46.36 & 30.49 & 17.05 & 38.53 & 27.47 & 16.24 & 33.35 \\
                             & 5 & 72.44 & 55.76 & 34.61 & 61.68 & 50.19 & 32.81 & 50.57 & 47.12 & 30.88 & 17.64 & 38.96 & 27.77 & 16.77 & 34.18 \\
                             & 6 & 70.56 & 55.84 & 33.85 & 60.67 & 50.31 & 31.99 & 50.22 & 46.49 & 30.10 & 17.05 & 38.36 & 27.23 & 16.26 & 33.40 \\
        \specialrule{.1em}{.05em}{.05em}
	\end{tabular}
	\caption{\small The performance (\%) of VSLNet with data debiasing (DD) strategy over different number of clip $N_{clip}$ on the Charades-CD and ActivityNet-CD datasets.}
	\label{tab:full_n_clip_analysis}
\end{table*}

\begin{table*}[htbp!]
    \small
    \centering
    \setlength{\tabcolsep}{3 pt}
	\begin{tabular}{c | c | c c c | c c c | c | c c c | c c c | c}
		\specialrule{.1em}{.05em}{.05em}
		\multirow{3}{*}{Split} & \multirow{3}{*}{Method} & \multicolumn{7}{c |}{Charades-CD} & \multicolumn{7}{c}{ActivityNet-CD} \\
		\cline{3-16}
		& & \multicolumn{3}{c |}{$\text{R@}1,\text{IoU@}\mu$} & \multicolumn{3}{c |}{$\text{dR@}1,\text{IoU@}\mu$} & \multirow{2}{*}{mIoU} & \multicolumn{3}{c |}{$\text{dR@}1,\text{IoU@}\mu$} & \multicolumn{3}{c |}{$\text{dR@}1,\text{IoU@}\mu$} & \multirow{2}{*}{mIoU} \\
        & & $\mu$=0.3 & $\mu$=0.5 & $\mu$=0.7 & $\mu$=0.3 & $\mu$=0.5 & $\mu$=0.7 & & $\mu$=0.3 & $\mu$=0.5 & $\mu$=0.7 & $\mu$=0.3 & $\mu$=0.5 & $\mu$=0.7 & \\
        \specialrule{.1em}{.05em}{.05em}
        \multirow{3}{*}{iid} & VSLNet & 75.09 & 60.51 & 41.07 & 66.72 & 56.12 & 39.25 & 54.39 & 63.18 & 49.16 & 32.37 & 52.41 & 43.40 & 31.37 & 47.57 \\
                             & +V-MD  & 73.63 & 60.87 & 39.61 & 65.68 & 56.41 & 38.04 & 54.11 & 61.27 & 46.88 & 30.16 & 51.80 & 41.94 & 28.41 & 45.09 \\
                             & +Q-MD  & 73.51 & 60.39 & 40.83 & 65.56 & 55.95 & 38.95 & 53.41 & 60.88 & 46.14 & 30.94 & 51.67 & 41.37 & 29.15 & 45.00 \\
                             & +MD    & 73.39 & 59.17 & 39.13 & 65.22 & 54.69 & 37.41 & 53.26 & 58.73 & 45.10 & 29.32 & 50.23 & 40.51 & 27.60 & 43.62 \\
        \hline
        \multirow{3}{*}{ood} & VSLNet & 65.93 & 48.18 & 28.89 & 55.85 & 43.29 & 27.20 & 45.56 & 41.71 & 23.31 & 12.06 & 32.99 & 20.96 & 11.59 & 29.95 \\
                             & +V-MD  & 68.24 & 52.74 & 32.47 & 58.21 & 47.50 & 30.53 & 48.42 & 42.85 & 27.23 & 14.56 & 36.07 & 24.91 & 13.96 & 30.45 \\
                             & +Q-MD  & 67.11 & 51.59 & 32.18 & 57.90 & 46.63 & 30.33 & 47.60 & 42.89 & 27.06 & 14.60 & 36.14 & 24.73 & 14.01 & 30.58 \\
                             & +MD    & 70.79 & 54.90 & 33.30 & 60.59 & 49.26 & 31.64 & 49.49 & 42.43 & 27.70 & 15.01 & 36.08 & 25.33 & 14.41 & 30.55 \\
        \hline
        \multirow{3}{*}{all} & VSLNet & 67.72 & 50.60 & 31.28 & 57.98 & 45.81 & 29.56 & 47.30 & 46.02 & 28.49 & 16.14 & 36.89 & 25.47 & 15.66 & 33.19 \\
                             & +V-MD  & 69.92 & 54.34 & 33.87 & 59.67 & 49.25 & 32.00 & 49.53 & 46.54 & 31.18 & 17.12 & 39.23 & 28.13 & 16.75 & 33.35 \\
                             & +Q-MD  & 68.37 & 53.31 & 33.87 & 59.40 & 48.45 & 32.02 & 48.74 & 46.50 & 30.88 & 17.28 & 39.26 & 27.86 & 16.85 & 33.47 \\
                             & +MD    & 70.25 & 55.79 & 35.30 & 60.93 & 50.63 & 33.33 & 50.40 & 45.70 & 31.19 & 17.88 & 38.92 & 28.38 & 17.05 & 33.17 \\
        \specialrule{.1em}{.05em}{.05em}
	\end{tabular}
	\caption{\small The performance (\%) of VSLNet with different model debiasing (MD) strategies on the Charades-CD and ActivityNet-CD datasets.}
	\label{tab:full_md_analysis}
\end{table*}

\end{document}